\documentclass[12pt]{article}

\usepackage{scicite}

\usepackage{times}
\usepackage{xspace}
\usepackage{color}
\usepackage[utf8]{inputenc} 
\usepackage[T1]{fontenc}    
\usepackage[hidelinks]{hyperref}       
\usepackage{url}            
\usepackage{adjustbox}
\usepackage{graphicx}
\usepackage[usenames,dvipsnames]{pstricks}
\usepackage{epsfig}
\usepackage{ifxetex}
\usepackage{pifont}
\usepackage{amsmath}
\usepackage{amsfonts}
\usepackage{enumitem}
\usepackage{tikz}

\newenvironment{sciabstract}{%
\begin{quote} \bf}
{\end{quote}}

\ifxetex
	\newcommand{\xefig}[1]{#1}
\else
	\newcommand{\xefig}[1]{}
\fi

\newcommand{\Ent}[0]{\ensuremath{{\cal E}}\xspace}
\newcommand{\Rel}[0]{\ensuremath{{\cal R}}\xspace}

\newcommand{\KGCard}[0]{\ensuremath{\ell}\xspace}
\newcommand{\EntCard}[0]{\ensuremath{m}\xspace}
\newcommand{\RelCard}[0]{\ensuremath{n}\xspace}
\newcommand{\ent}[0]{\ensuremath{e}\xspace}
\newcommand{\rel}[0]{\ensuremath{r}\xspace}
\newcommand{\sub}[0]{\ensuremath{s}\xspace}
\newcommand{\obj}[0]{\ensuremath{o}\xspace}

\newcommand{\EntCardi}[0]{\ensuremath{\chi}\xspace}

\newcommand{\degree}[0]{\ensuremath{k}\xspace}
\newcommand{\degreeProb}[0]{\ensuremath{P(k)}\xspace}

\newcommand{\newProb}[0]{\ensuremath{\sigma}\xspace}

\newcommand{\maxDegree}[0]{\ensuremath{\degree_{\max}}\xspace}
\newcommand{\propProb}[0]{\ensuremath{\rho_\rel}\xspace}
\newcommand{\attachProb}[0]{\ensuremath{\beta_\rel}\xspace}
\newcommand{\attachExp}[0]{\ensuremath{\alpha_\rel}\xspace}

\newcommand{\propProbC}[0]{\ensuremath{\rho}\xspace}
\newcommand{\attachProbC}[0]{\ensuremath{\beta}\xspace}
\newcommand{\attachExpC}[0]{\ensuremath{\alpha}\xspace}

\newcommand{\mean}[1]{\ensuremath{\mathbb{E}\left[{#1}\right]}\xspace}
\newcommand{\variance}[1]{\ensuremath{\mathbb{V}\left({#1}\right)}\xspace}
\newcommand{\Prob}[1]{\ensuremath{\mathbb{P}\left({#1}\right)}\xspace}
\renewcommand{\aa}[0]{\ensuremath{a}\xspace}
\newcommand{\bb}[0]{\ensuremath{b}\xspace}
\newcommand{\cc}[0]{\ensuremath{c}\xspace}

\newcommand{\uri}[1]{\ensuremath{\small \textrm{\textsf{#1}}}\xspace}

\newcommand{\graph}[0]{\ensuremath{G}\xspace}

\title{The Structure and Dynamics of Knowledge Graphs, with Superficiality}

\author
{Loïck Lhote$^{1}$, Béatrice Markhoff$^{2}$, Arnaud Soulet$^{3\ast}$\\
\\
\normalsize{$^{1}$GREYC, Normandy University}\\
\normalsize{Boulevard Maréchal juin, Caen, F-14032, France}\\
\normalsize{$^{2}$CITERES, University of Tours}\\
\normalsize{$^{3}$LIFAT, University of Tours}\\
\normalsize{3 place Jean Jaurès, Blois, F-41000, France}\\
\\
\normalsize{$^\ast$To whom correspondence should be addressed; E-mail:  arnaud.soulet@univ-tours.fr.}
}

\date{}

\begin{document}


\baselineskip24pt


\maketitle 

\vspace{-1cm}
 \begin{tikzpicture}[remember picture,overlay]
 \hypersetup{hidelinks}
 \node[anchor=north west,yshift=360pt,xshift=0pt]
 { \href{https://doi.org/10.24072/pci.networksci.100113}{\includegraphics[height=35mm]{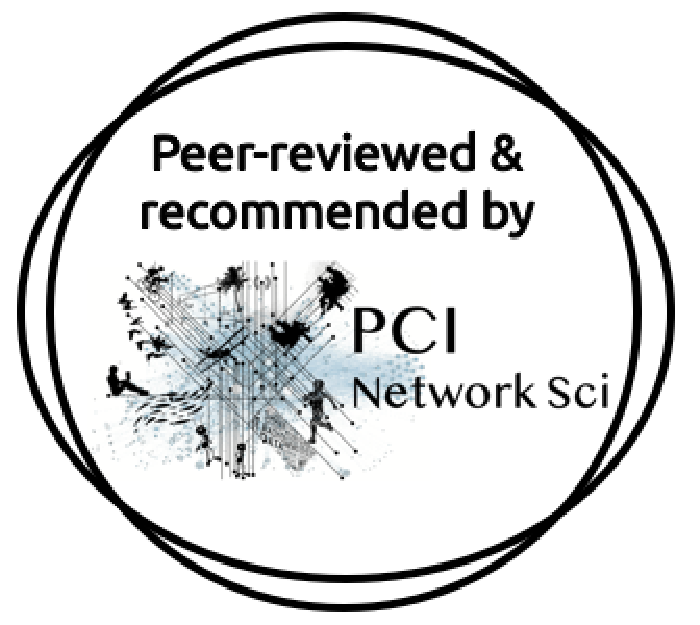}} } ;
 \end{tikzpicture}


\begin{sloppypar}
\begin{sciabstract}
    Large knowledge graphs combine human knowledge garnered from projects ranging from academia and institutions to enterprises and crowdsourcing. 
    Within such graphs, each relationship between two nodes represents a basic fact involving these two entities. 
    The diversity of the semantics of relationships constitutes the richness of knowledge graphs, leading to the emergence of singular topologies, sometimes chaotic in appearance. However, this complex characteristic can be modeled in a simple way by introducing the concept of superficiality, which controls the overlap between relationships whose facts are generated independently. With this model, superficiality also regulates the balance of the global distribution of knowledge by determining the proportion of misdescribed entities. This is the first model for the structure and dynamics of knowledge graphs. It leads to a better understanding of formal knowledge acquisition and organization.
\end{sciabstract}
\end{sloppypar}



\section*{Knowledge Graphs}

\begin{figure}
\begin{center}
\includegraphics{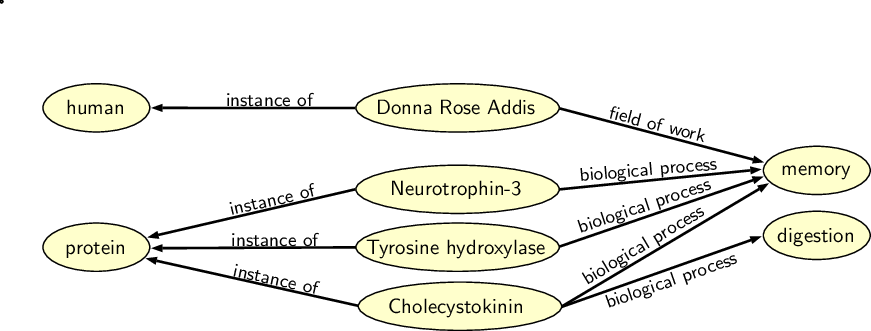}
\caption{
\emph{A very small sample of the Wikidata knowledge graph.} 
{\small
The yellow nodes represent entities (like \uri{protein}, \uri{Neurotrophin-3} or \uri{memory}) and the arrows represent relationships (like \uri{instance of}, \uri{field of work} or \uri{biological process}). A fact corresponds to a subject entity linked to an object entity (e.g., $\uri{(Neurotrophin-3, biological process, memory)}$). Note that in the complete knowledge graph, a node often has both incoming and outgoing links.
}}
\label{fig:ToyKG}
\end{center}
\end{figure}

\begin{sloppypar}
A knowledge graph is a knowledge base represented as a directed graph whose vertices are the entities and whose labeled edges are their relationships. Each edge represents a fact similar to an elementary sentence that relates a subject to an object. For example, in Figure~\ref{fig:ToyKG}, the fact $\uri{(Neurotrophin-3, biological process, memory)}$ represents the involvement of the protein Neurotrophin-3 (subject entity) in the biological process (relationship) of memory (object entity).
Since the development of the Semantic Web \cite{berners2001semantic}, knowledge graphs are often associated with open data projects of the Web of data. This movement has led to the development of scientific and institutional knowledge graphs of unprecedented size, notably in cultural heritage \cite{Bikakis2021} and life sciences \cite{kamdar2021empirical,santos2022knowledge}. At the same time, other projects aim to build encyclopedic knowledge graphs such as Yago \cite{suchanek2008yago}, DBpedia \cite{lehmann2015dbpedia} or Wikidata \cite{vrandevcic2014wikidata}, whose collaborative editing has made it possible to group more than 14 billion facts.
For the Wikidata relationship \uri{biological process} alone, there are more than 1.1M stated facts, comparable to the triplet $\uri{(Neurotrophin-3, biological process, memory)}$.
%
%
This large semantically rich data source allows for creating new scientific hypotheses by cross-referencing knowledge, manually or through machine learning.
In this context, understanding the topology of knowledge graphs is fundamental to estimate how complete the accumulated knowledge is and predicting their evolution.
%
Only this understanding can guarantee that  new knowledge induced from these knowledge graphs, manually or through machine learning, is valid with respect to the real world.
\end{sloppypar}

The topology of knowledge graphs and their evolution remains largely unknown because of the complex interactions between relationships.
In network science, it is well known that the preferential attachment mechanism \cite{barabasi1999emergence} plays a central role in constructing graphs. However, the proposed models apply to most of the networks where all the links obey the same preferential attachment and, in some cases, to graphs with two kinds of relationships, whereas a knowledge graph contains dozens or even hundreds of relationships. 
In this paper, we show how the multiplexing of different kinds of relationships (as considered in \cite{de2013mathematical,boccaletti2014structure}) brings out complex and unexpected topological phenomena.
By exploring knowledge graphs representing diverse knowledge, such as documentary heritage, bioactive molecules \cite{gaulton2012chembl} or encyclopedic knowledge \cite{vrandevcic2014wikidata}, we observe that their topology does not boil down to a simple power law like simplex networks, such as the Web or citation graphs \cite{newman2005power}.
%
Naturally, the superposition of the singular dynamics of each relationship leads to a multimodal probability distribution.
%
More surprisingly, the multiplexing of relationships also creates strong irregularities in the probability distribution, especially for the outgoing connectivity of entities. This phenomenon is crucial because it concerns the entities with a low connectivity, corresponding to the majority of the knowledge graph's entities.
To understand the origin of this phenomenon, we introduce the notion of superficiality, that is the probability of adding a new entity in the graph when a relationship must be completed with an additional entity (otherwise the relationship is completed with an entity already existing in the graph). The superficiality controls the concentration of kinds of relationships per entity, involving the presence or absence of these perturbations and determining the proportion of misdescribed entities.

\section*{Preferential Attachment and Multiplex Networks}

\begin{sloppypar}

Preferential attachment, defined by Barabási and Albert \cite{barabasi1999emergence}, is one of the main mechanisms to explain the emergence of network structure.
It consists in adding links giving a priority to nodes that already have strong connectivity. More precisely, the probability of adding a new link connecting vertex $i$ will be proportional to $k_i^\alpha$ with $k_i$ being the degree of $i$ (its connectivity) and $\alpha>0$ being an important parameter to consider the diversity of observed distributions \cite{Broido2019}. A linear exponent $\alpha = 1$ infers a graph whose degrees follow a power law. The proportion of nodes of degree $k$ is proportional to $k^{-\gamma }$ for $k$ large.
For a sublinear exponent $0 < \alpha < 1$, the degree distribution tends to a stretched exponential function $e^{-\degree \gamma}$.
Several proposals for generalizing the Barabási-Albert model are also relevant in the context of modeling knowledge graphs.
As in directed networks, it is reasonable to distinguish between the preferential attachment of incoming connectivity and outgoing connectivity \cite{bollobas2003directed}. For example, the Wikidata relationship \uri{biological process} describing the involvement of entities (e.g., proteins or genes) in biological processes (e.g., memory or digestion) is asymmetric. The entities involved are unevenly distributed between the different processes with only 1 fact for more than 15 thousand biological processes against several tens of thousands of facts for the main biological processes such as metabolism or regulation of DNA-induced transcription.
This corresponds to a quasi-linear preferential attachment (i.e., $\alpha \approx 1$) while each entity is involved in a close number of processes reflecting a weak preferential attachment (i.e., $\alpha \approx 0$). 
It is thus necessary to take into account sublinear attachments for some relationships \cite{krapivsky2000connectivity}. 
A major drawback of these models and more elaborate ones \cite{chung2006complex,giroire2023random} dedicated to simplex networks is that they infer laws where multimodality or localized drops in connectivity as observed in knowledge graphs (pointed by the red arrows in the graphs of Figure~\ref{fig:KGs}) can not be explained. In fact, a knowledge graph is similar to a multiplex network (very similar with multi-layer or multi-dimensional networks) where each relationship constitutes a distinct layer \cite{de2013mathematical,boccaletti2014structure}. The study of generative models for multiplex networks is a tough and little studied scientific challenge, which explains why our proposal is the first generative model adapted to knowledge graphs. 
To the best of our knowledge, there are only generative models for duplex networks \cite{nicosia2013growing,kim2013coevolution} that consider correlations between two layers. Their analytical results would be difficult to generalize to knowledge graphs which have many relationships. For instance, Wikidata had about 1,500 relationships in 2022. Moreover, this would mean defining at least a quadratic number of parameters with the number of relationships. Instead of considering strong interactions between a small number of layers, our proposal aims at privileging weak interactions between a large number of layers.
\end{sloppypar}

\section*{Generative Model with Superficiality}

\subsection*{Description of the model}

In our model, the only link between the different relationships is the sharing of entities which varies according to the level of superficiality. 
We generate the facts of each relationship separately and independently by distributing them over shared entities. In this way, we take into account the semantics of each relationship by distinguishing its sets of subjects and objects, and by assigning it a specific preferential attachment.

\begin{figure}
\begin{center}
\includegraphics[scale=0.85, keepaspectratio]{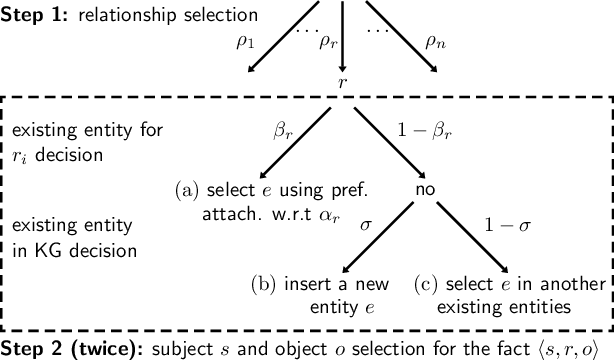}
\hspace*{1cm}
\resizebox{0.35\textwidth}{!}{
\raisebox{1.00\height}{
{\small
\begin{tabular}{|c|l|}
\hline
\multicolumn{2}{|c|}{Global KG parameters} \\
\hline
\KGCard & number of facts \\
\EntCard & number of entities \\
\newProb & superficiality \\
\hline
\hline
\multicolumn{2}{|c|}{Parameters for each relationship \rel} \\
\hline
\propProb & proportion of facts  \\
\attachExp & attachment exponent  \\
\attachProb & attachment probability  \\
\hline
\multicolumn{2}{c}{} \\
\hline
\multicolumn{2}{|c|}{Studied connectivities for entities} \\
\hline
$P_{\attachExpC_r,\attachProbC_r}(\degree)$ & connectivity for \rel per entity \\ 
$P(k)$ & total connectivity per entity \\ 
$P(r)$ & distinct relationships per entity \\ 
\hline
\end{tabular}
}
}
}
\end{center}
\caption{
\emph{Generative model of a fact $\langle s, r_i, o \rangle$.} 
{\small
The first step consists in randomly drawing the relationship of the fact proportionally to \propProb. Then, the subject \sub and the object \obj are chosen by applying the boxed procedure. For each entity \sub or \obj, there are 3 cases: (a) choose the entity with a preferential attachment mechanism (with a probability \attachProb), (b) insert a new entity (with a probability $(1- \attachProb) \times \newProb$) or (c) choose an entity from the knowledge graph already described by another relationship (with a probability $(1- \attachProb) \times (1 - \newProb$)).
At initialization, it is necessary to force the insertion of an entity for the relationship \rel to allow the case (a) and the insertion of new entities in the knowledge graph to allow the case (c).}}
\label{fig:Process}
\end{figure}

Precisely, we add at each time a new fact $(\sub, \rel, \obj)$ in the knowledge graph by starting to randomly choose a relationship $\rel$ with a probability proportional to \propProb (Step 1 of Figure~\ref{fig:Process}).
The subject and object entities of this new fact are then drawn according to the same strategy (Step 2 repeated for the choice of subject \sub and object \obj):
\begin{enumerate}[label=(\alph*)]
%
\item with probability \attachProb, we focus on the semantics  by randomly drawing an entity \ent with a probability proportional to $\degree_{\ent,\rel}^{\attachExp}$ where $\degree_{\ent,\rel}$ is the degree of \ent for the relationship \rel; 
\item with a probability (1-\attachProb)\newProb, we add a new entity that does not yet belong to the knowledge graph; and 
\item with a probability $(1-\attachProb)(1- \newProb)$, we uniformly draw an entity among those already existing in the knowledge graph because added for completing another relationship.
\end{enumerate}
In this model, the superficiality \newProb controls the probability to create a new entity or to choose an existing entity in the neighborhood of an other relationship  when it is necessary to add an entity that is not yet described by the facts of the relationship.
We will see later that the higher this parameter \newProb, the more the facts added to the knowledge graph are spread over numerous entities, leading on average to a \emph{superficial} knowledge of each entity.


\subsection*{How to compute model parameters?}

In practice, we can parameterize our generative model to verify its ability to reproduce existing knowledge graphs.
Each probability \propProb corresponds to the proportion of facts of the relationship \rel among all the facts of the knowledge graph.
For a relationship \rel, if the generation process is repeated $n_{\rel}$ times, the average number of distinct entities generated for this relationship is then $(1- \attachProb) \times n_{\rel}$. Thus, for a given knowledge graph, \attachProb is easily parameterized based on the number of distinct entities in the relationship \rel and its number of facts $n_{\rel}$.
%
The superficiality parameter \newProb is then derived by considering both the number of facts \KGCard and the number of entities \EntCard present in the knowledge graph. Indeed, the average number of distinct entities generated by our model by repeating it \KGCard times will be $\sum_{\rel \in \Rel} \propProb \times (1 - \attachProb) \times \newProb \times \KGCard$. To obtain \EntCard entities on average, the probability \newProb must be equal to $\EntCard / (\sum_{\rel \in \Rel} \propProb \times (1 - \attachProb) \times \KGCard)$.
With the number of entities in the numerator and the number of facts in the denominator, the superficiality \newProb might look like a kind of an average number of facts per entity normalized between 0 and 1. But more insights are captured by weighting the number of facts with the proportion and probability of attachment.

\subsection*{Comparisons with real knowledge graphs}

Our random generative model reproduces well the general shape of the distribution of incoming and outgoing degrees measured in the three major knowledge graphs (Figure~\ref{fig:KGs}). In particular, the curvatures induced by the multimodality of our model are visible and close to those observed in real data (e.g., for Wikidata outgoing connectivity). For outgoing connectivity, the drop in the proportion \degreeProb of entities of degree $k$ for low degrees $k$ is perfectly transcribed, especially on Wikidata where the probability drops for $k=1$. 
However, our random model misses some of the micro-variations, especially for the outgoing connectivities of Chemical Biology Database (ChEMBL)\footnote{The database is freely available and maintained by the European Bioinformatics Institute (EBI), of the European Molecular Biology Laboratory (EMBL).}. 
For this graph, as the number of relationships is limited (only 50), localized perturbations due to inter-relationship correlations ignored by our model are more visible.
However, except for the outgoing connectivity in ChEMBL, the Kullback-Leibler divergence is small for all distributions indicating high proximity between the real and generated distributions.  
We also observe for these three knowledge graphs that the superficiality is lower for the outgoing connectivity where precisely, the variations of the proportion \degreeProb are the most chaotic.
We have conducted an ablation study to compare the generative model that we define 
to the generative models of Barab\'asi-Albert \cite{barabasi1999emergence}, and Bollob\'as \cite{bollobas2003directed} (Figure~\ref{fig:CompKGs}).
 First, we observed that the parameterization of the preferential attachment is always beneficial. Our second observation is that the use of a multiplex model is relevant. Of course, the simplex approaches do not take into account the oscillations, contrary to our generative model. 
Simplex models poorly reproduce entities with high degrees by generating entities with far too many facts (with one or two orders of magnitude higher than expected, like for the out-degree of BnF and Wikidata).
 Finally, on the three knowledge graphs used for experiments (BnF, ChEMBL and Wikidata), it is clear that our model is the most satisfactory. This study and its results are detailed in the supplementary materials. 

\subsection*{Degree distributions for a given relationship }
At the level of a relationship, the parameters \attachExp and \attachProb are intrinsic characteristics of the relationship \rel. The exponent \attachExp constrains the distribution of facts between a uniform drawing of entities ($\attachExp = 0$) and a high preferential attachment ($\attachExp = 1$). The probability \attachProb regulates the average number of facts per entity corresponding to this relationship. The fact acquisition rate for the relationship \rel describing the entity \ent is as follows: $\degree_{\ent,\rel}(t + 1) - \degree_{\ent,\rel}(t) = \attachProb \times {\degree_{\ent,\rel}(t)^{\attachExp}}/{Z_\rel(t)}$ with the normalization $Z_\rel(t) = \sum_{\ent \in \Ent} \degree_{\ent,\rel}(t)^{\attachExp}$.  
Here $t$ and $t+1$ are two successive distinct moments where the relationship $r$ is chosen after the creation of $e$.
For a linear exponent $\attachExp = 1$, we find a variant of Barab\'asi-Albert model \cite{barabasi1999emergence} because $Z_\rel(t) = t$. Following the same method of resolution, it is possible to compute the probability distribution which follows a power law of exponent $1 + 1 / \attachProb$.
For a sublinear exponent $\attachExp < 1$, our model is close to the one studied in \cite{krapivsky2000connectivity} with the sublinear connection kernel leading only to explicit results for some limit cases. Nevertheless, whatever the exponent \attachExp and the probability \attachProb, it is clear that the average proportion  of entities attached to $\degree$ facts involving $\rel$, noted $P_{\attachExp,\attachProb}(\degree)$, is always decreasing with the degree \degree which does not explain the strong irregularities observed in many knowledge graphs.

\subsection*{Degree distributions in the whole graph}

The average proportion of entities having total degree \degree (or equivalently involved in \degree facts), is called the total connectivity \degreeProb.
The variations observed in \degreeProb comes from the combined effect of the average proportions $P_{\attachExpC_r,\attachProbC_r}(\degree)$ of the different relationships \rel.
For example, the incoming connectivity of the entity \uri{memory} aggregates both the facts of the relationship \uri{biological process}, but also those of other relationships such as the relationship \uri{field of work}, indicating people and institutions working on memory. 
The multimodality of \degreeProb is simply explained by the superposition of these different average proportions. In contrast, the observed drops are related to the distribution of the number of distinct relationships for all entities. In the case where a majority of entities would be described by two kinds of relationships (say \uri{biological process} and \uri{field of work}), the number of entities with only one fact would become lower than the number of entities with 2 facts inducing $P(1) < P(2)$.

\begin{figure}
\begin{center}
\includegraphics[angle=-90, scale=0.375, keepaspectratio]{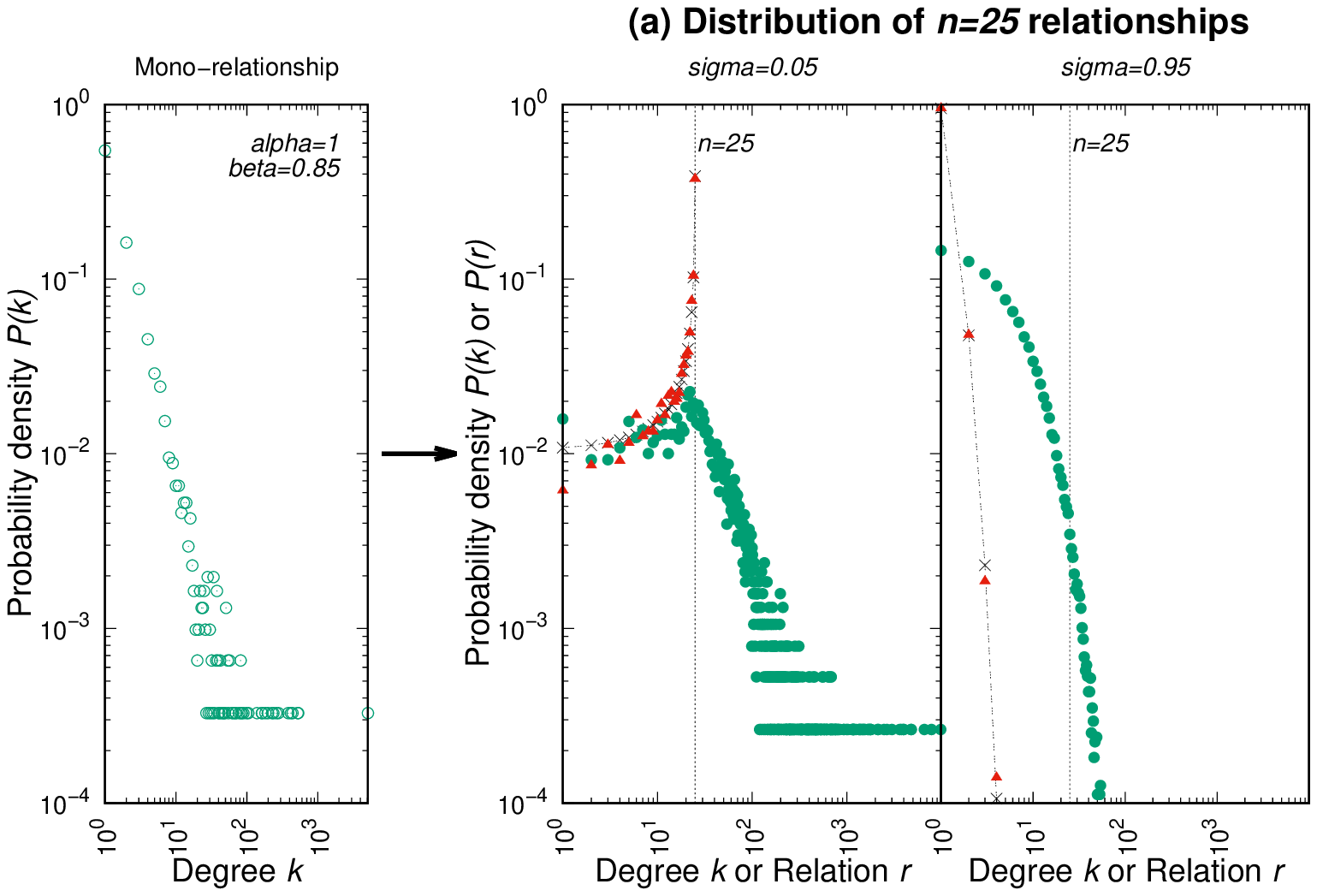}
\includegraphics[angle=180, scale=0.275, keepaspectratio]{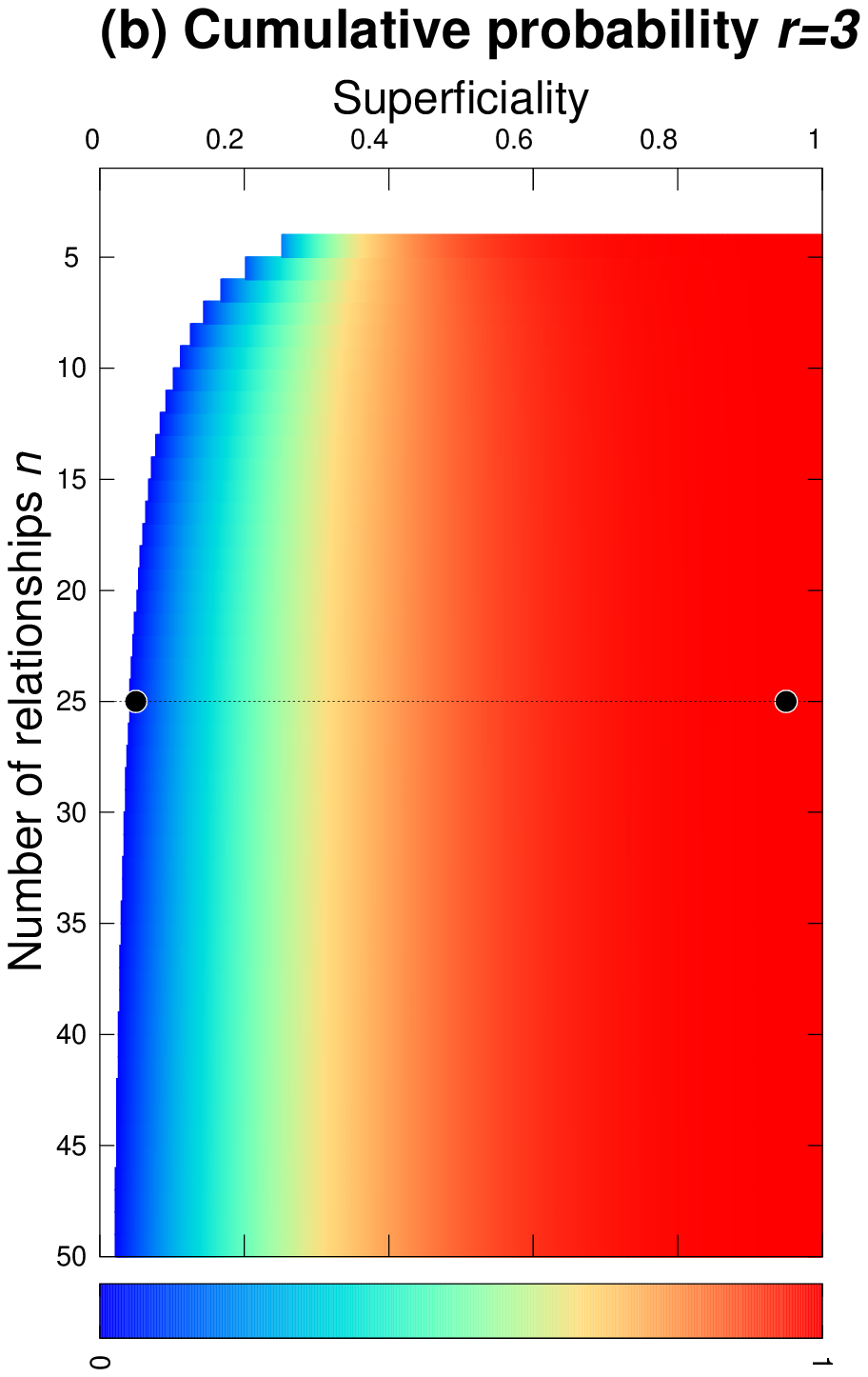}
\caption{
\emph{Simulation of the superficiality impact on the proportion of misdescribed entities.} 
{\small
The left plot reports the probability density of one relationship with $\attachProbC = 0.85$ and $\attachExpC = 1$ ({\tiny \textcolor{teal}{\ding{109}}}). On the middle, the multiplexing of 25 relationships with these same characteristics is plotted for two extreme superficialities ({\tiny \textcolor{teal}{\ding{108}}}). For a low superficiality $\newProb = 0.05$, we observe a huge drop of the probability $P(k)$ between $k=1$ and 25 due to an increasing probability density $P(r)$ ({\tiny \textcolor{red}{\ding{115}}}), compliant to Equation~\ref{equ:Probability} ({\tiny $\times$}). For $\newProb = 0.95$, the probability density is much more regular because both distributions $P(k)$ and $P(r)$ are strictly decreasing.
On the right-hand side, the proportion of misdescribed entities $P(\rel_\ent(t) \le 3)$ of the knowledge graph is very high whenever $\newProb \ge 0.5$ (red area) meaning that most of the entities are described by less than 3 relationships. This proportion converges to $1 - (1 - \newProb)^3$ when the number of relationships tends to infinity. 
Note that the two points correspond to the projections of the two simulations (a).
}} 
\label{fig:Simulation}
\end{center}
\end{figure}

To theoretically analyze this multiplexing phenomenon, we now consider that all the relationships have equivalent dynamics with the same proportion \propProbC, the same attachment probability \attachProbC and the same exponent \attachExpC.
It is possible to calculate the average proportion $P(\rel_\ent(t) = r)$ (or simply, $P(r)$) that an entity has $r$ distinct relationships by observing that the evolution of $\rel_\ent(t)$ depends empirically on two terms (Step 2c of Figure~\ref{fig:Process}). 
First, the probability of adding a new relationship $\rel$ to an entity $\ent$ (already drawn) at time $t+1$ is $(\RelCard -\rel_\ent(t) ) \times \propProbC \times (1 - \attachProbC) \times (1 - \newProb)$. 
Secondly, only an entity created by another relationship than $\rel$ can be chosen to receive a new fact for $\rel$. At time $t$, there are $t \times (\RelCard - 1) \times \propProbC \times (1 - \attachProbC) \times \newProb$ entities which were added in the graph by a relationship to which it is necessary to withdraw those added by the relationship $\rel$ namely $t \times \propProbC \times (1 - \attachProbC) \times (1 - \newProb)$.  The probability of choosing a particular entity is therefore $1 / \left[ \propProbC \times (1 - \attachProbC) \times (\newProb \RelCard - 1 ) \times t \right]$. For this probability to make sense, the generation of a knowledge graph for a superficiality \newProb imposes a minimal number of relationships: $\RelCard > 1 / \newProb - 1$. 
Considering these two probabilities, the acquisition rate of a new relationship is the following after simplification (for $\RelCard > 1 / \newProb - 1$):
\begin{equation}
\label{equ:Probability}
P(\rel_\ent(t) = r) = \frac{1 \times K_1 \times \dots \times K_{r-1}}{(1 + K_1) \times (1 + K_2) \times \dots \times (1 + K_r) }
\end{equation}
where $K_i = \frac{1 - \newProb}{\RelCard \newProb - {1}}\left(\RelCard-{i}\right)$.
The theoretical formula of Equation~\ref{equ:Probability} provides a valid justification for the observed perturbations for low degrees when the superficiality is low (in the case of outgoing connectivity). To illustrate this phenomenon, the simulation in Figure~\ref{fig:Simulation}(a) superimposes the $P(k)$ and $P(r)$ distributions. Of course, the average proportion of entities with $r$ relationships follows perfectly the theoretical values with an increase for $\newProb = 0.05$ and a decrease for $\newProb =  0.95$. The impact of $P(r)$ on $P(k)$ is particularly visible for the case where the superficiality is very low with $\newProb = 0.05$. In this case, the proportion of entities described by only a few facts, which indicates a form of ignorance, is much lower than when superficiality is high. 

\section*{Impact of Superficiality on Ignorance}

The quality of the knowledge about an entity is not only summarized by its number of facts because not all facts have the same value and some unrepresented ones can be inferred. Nevertheless, the presence of a large number of facts certainly reflects knowledge. With the open world assumption, it is more difficult to discern the entities for which there is no knowledge to represent and those for which the knowledge to represent is absent from the graph.
However, it is reasonable to think that each entity should be described by at least a few facts to specify its type and its basic interactions with other entities. For example, even a protein that is not involved in any biological process should be attached to the protein type, associated with the taxons where it exists, etc. 
Nevertheless, our model highlights a strong inequality in the distribution of knowledge, as for all systems based on a preferential attachment mechanism. Some entities accumulate more and more facts allowing them to refine their understanding, at the cost of introducing new entities linked to a few facts, thus maintaining a high proportion of misdescribed entities. 
Indeed, increasing the volume of knowledge does not globally reduce the level of ignorance underlying the graph, i.e. the proportion of entities that are described by a few facts $P(\degree_{\ent}(t) \le k)$ with $k$ small. Obviously, the acquisition over time of more facts does not modify this proportion conjugating the two distributions $P_{\attachExpC,\attachProbC}(\degree_{\ent}(t))$ and $P(\rel_{\ent}(t))$, which are independent of $t$.
More surprisingly, the increase of the number of relationships in the knowledge graph has a marginal effect on the level of ignorance through the proportion of entities described by a low number of relationships $P(\rel_{\ent}(t) \le r)$. We see that knowing more relationships changes this proportion of misdescribed entities to tend to a limit $\lim_{n \to +\infty} P(\rel_{\ent}(t) \le r) = 1-(1 - \newProb)^r$.
The proportion of misdescribed entities curiously increases with \RelCard until it reaches the limit $1-(1 - \newProb)^r$ when the superficiality \newProb is low. Although this increase is very slight, it reflects a form of knowledge paradox where the more we learn, the less we know.
On the contrary, if the superficiality \newProb is high, the proportion of misdescribed entities decreases with the addition of relationships to quickly reach the limit. This reduction is misleading since the final proportion converges toward a much higher ignorance than with a low \newProb. Reducing the proportion of misdescribed entities does not mean knowing well, but knowing better with the level of expertise set by \newProb. This observation should make one careful when exploiting knowledge graphs to avoid a kind of Dunning-Kruger effect \cite{kruger1999unskilled}. The fluctuation of the proportion of misdescribed entities does not allow to judge the quality of the knowledge graph; only the level of superficiality \newProb matters.
Figure~\ref{fig:Simulation}(b) plots $P(\rel_\ent(t) \le 3)$ according to the number of relationships \RelCard and the superficiality \newProb in knowledge graphs simulated with $\attachExp = 1$ and $\attachProb = 0.85$. With the color scale of the figure, an evolution from red to blue corresponds to a reduction of ignorance. Confirming the theoretical results, it appears clearly that the addition of relationships modifies little the level of ignorance. 
The only way to significantly reduce the proportion of misdescribed entities in a knowledge graph is to reduce its superficiality to concentrate more relationships on each entity. Yet, by repeating longitudinal measurements in the Wikidata knowledge graph between 2016 and 2022, we observed relative stability of superficiality (from 0.348 to 0.284 for outgoing connectivity and from 0.716 to 0.797 for ingoing connectivity). To take this further, it would be feasible to refine the analysis to identify parts of the graph with reasonable superficiality corresponding to the best-informed fields, from which one can safely induce new knowledge.

The superficiality is an essential parameter to properly model the construction of knowledge graphs, to understand their evolution and evaluate their quality.
In computer science, having a realistic theoretical model for knowledge graphs is crucial to optimize their interrogation by improving data storage and better estimate the cost of queries.
In knowledge engineering, our modeling is fundamental to estimate the robustness of the knowledge contained in these large graphs to induce new reliable knowledge, but also to study their vulnerability \cite{albert2000error}.
We believe that although simple, this description of knowledge graphs not only allows better exploitation of knowledge graphs in the Web of Data with possible opportunities in many domains, but it also opens the way to interdisciplinary research perspectives.
Finally, like Wikidata that is a mirror of the Wikipedia encyclopedia, each knowledge graph is a computerized representation of the knowledge of a field. Although this representation bias leads us to be cautious, our model could suggest assumptions about the organization of knowledge by considering our work as a form of computational epistemology.


\bibliography{superficiality}

\bibliographystyle{plain}

\section*{Acknowledgments}

The authors declare that they have received no specific funding for this study and that they have no financial or non-financial conflicts of interest.

\clearpage

\newcommand{\hide}[1]{#1}

\begin{figure}
\begin{center}
\hide{
\includegraphics[angle=-90, scale=0.33, keepaspectratio]{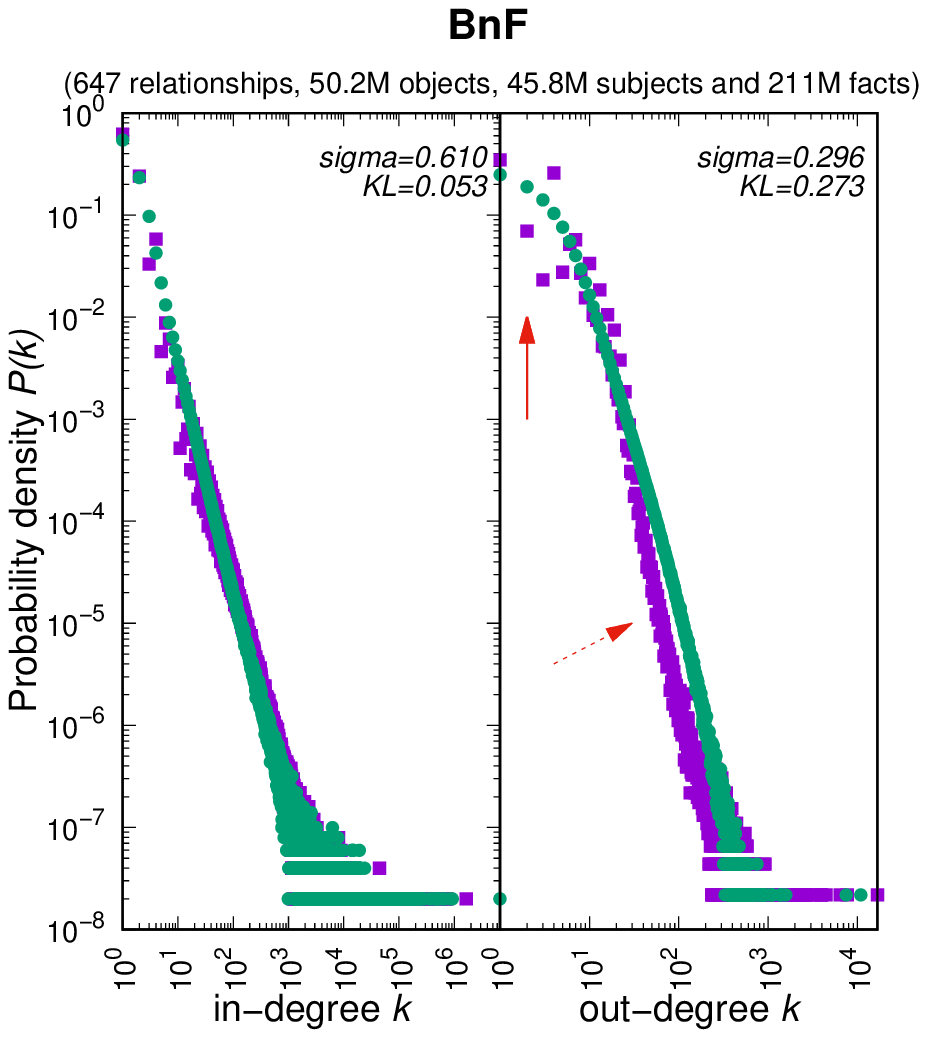}
\includegraphics[angle=-90, scale=0.33, keepaspectratio]{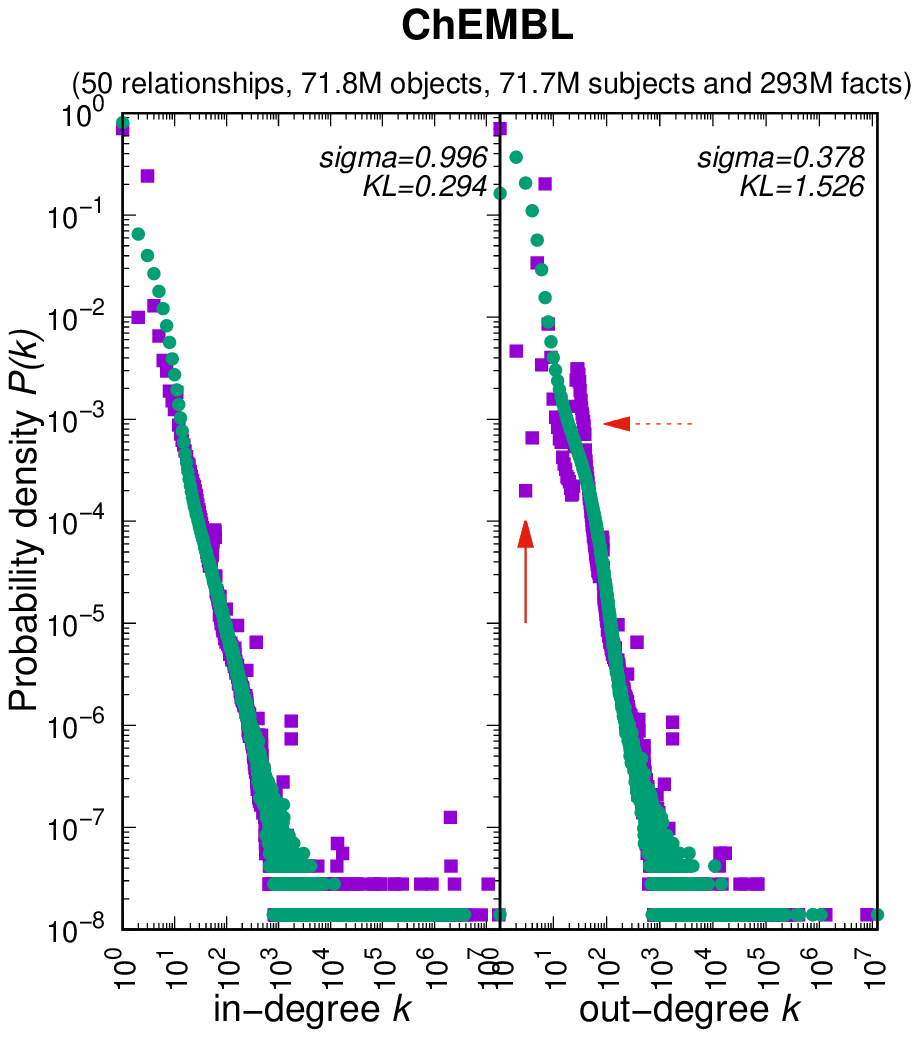}
\includegraphics[angle=-90, scale=0.33, keepaspectratio]{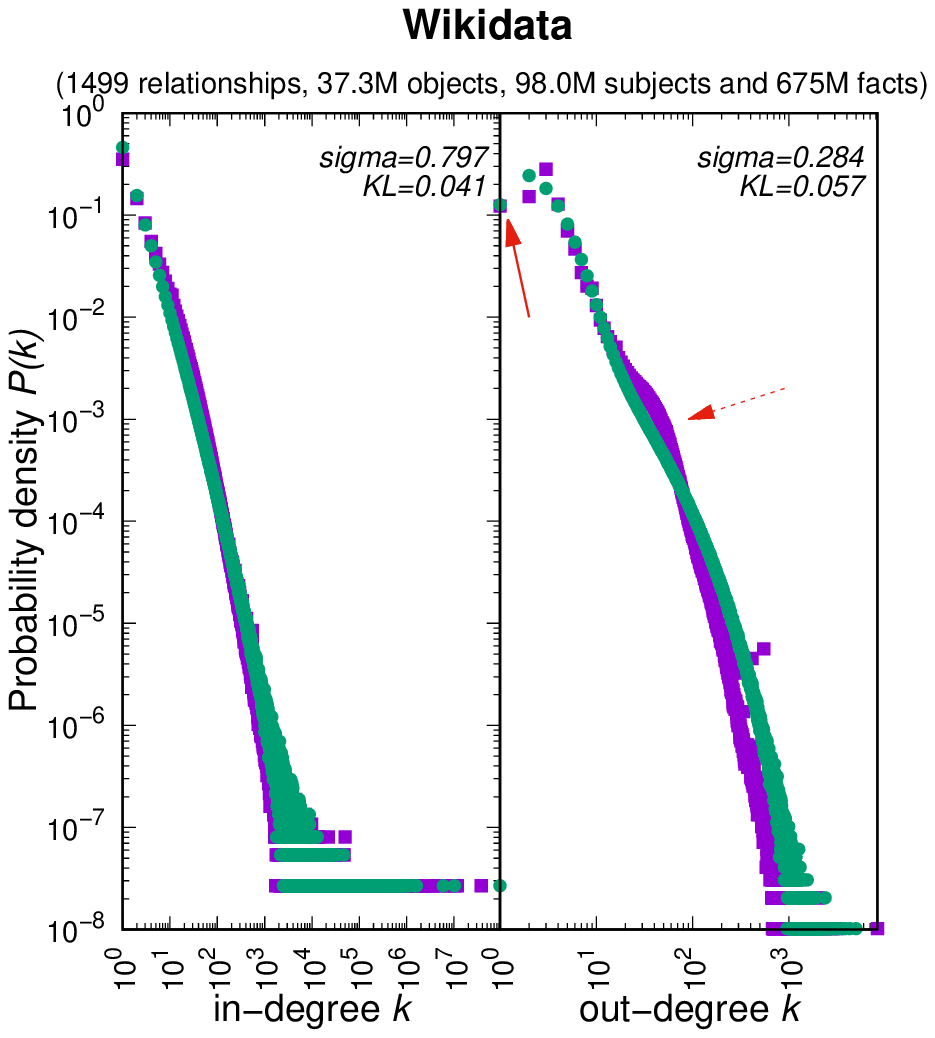}
}
\caption{
\emph{Comparison of real-world data ({\tiny \textcolor{violet}{\ding{110}}}) and data generated by our model ({\tiny \textcolor{teal}{\ding{108}}}) for BnF, ChEMBL and Wikidata with multiplexity phenomena highlighted by red arrows.} 
{\small
Dashed arrows point to phenomena explainable by multimodality while normal arrows point to unexpected drops of probability density. 
}}
\label{fig:KGs}
\end{center}
\end{figure}

\begin{figure}
\begin{center}
{\small
\hide{

(a) Barabási-Albert model

\includegraphics[angle=-90, scale=0.33, keepaspectratio]{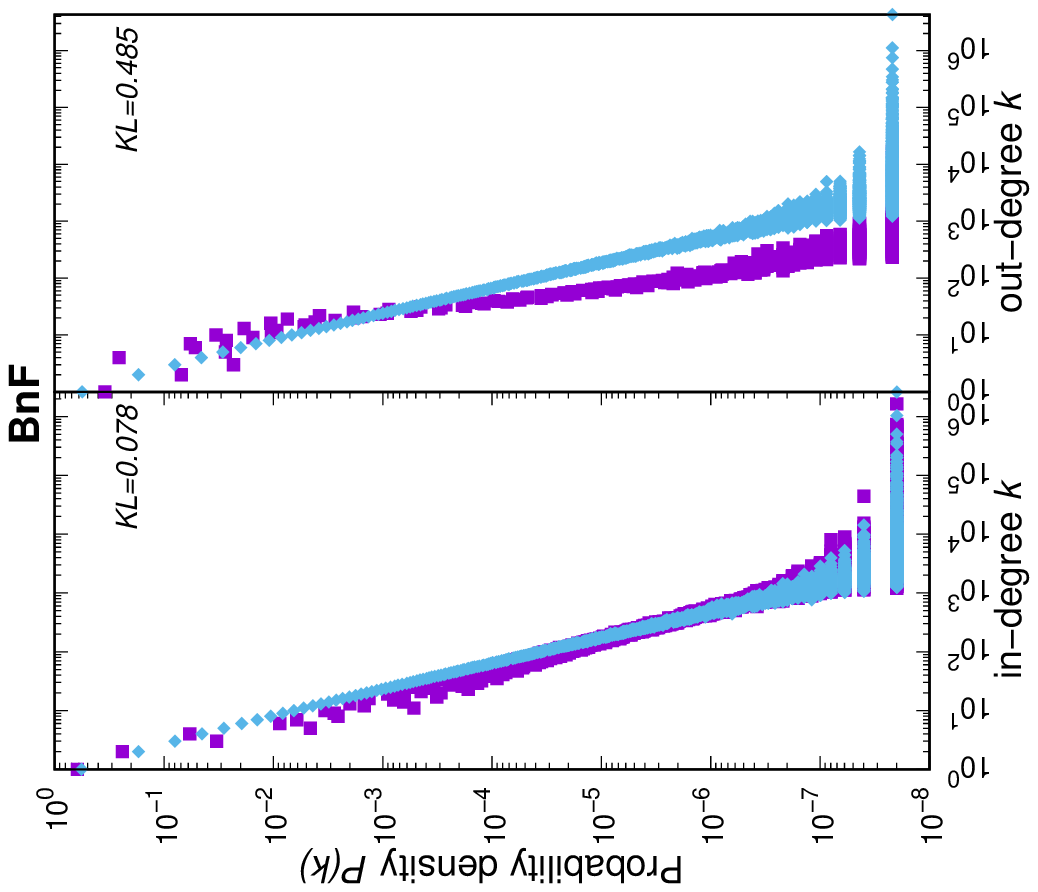}
\includegraphics[angle=-90, scale=0.33, keepaspectratio]{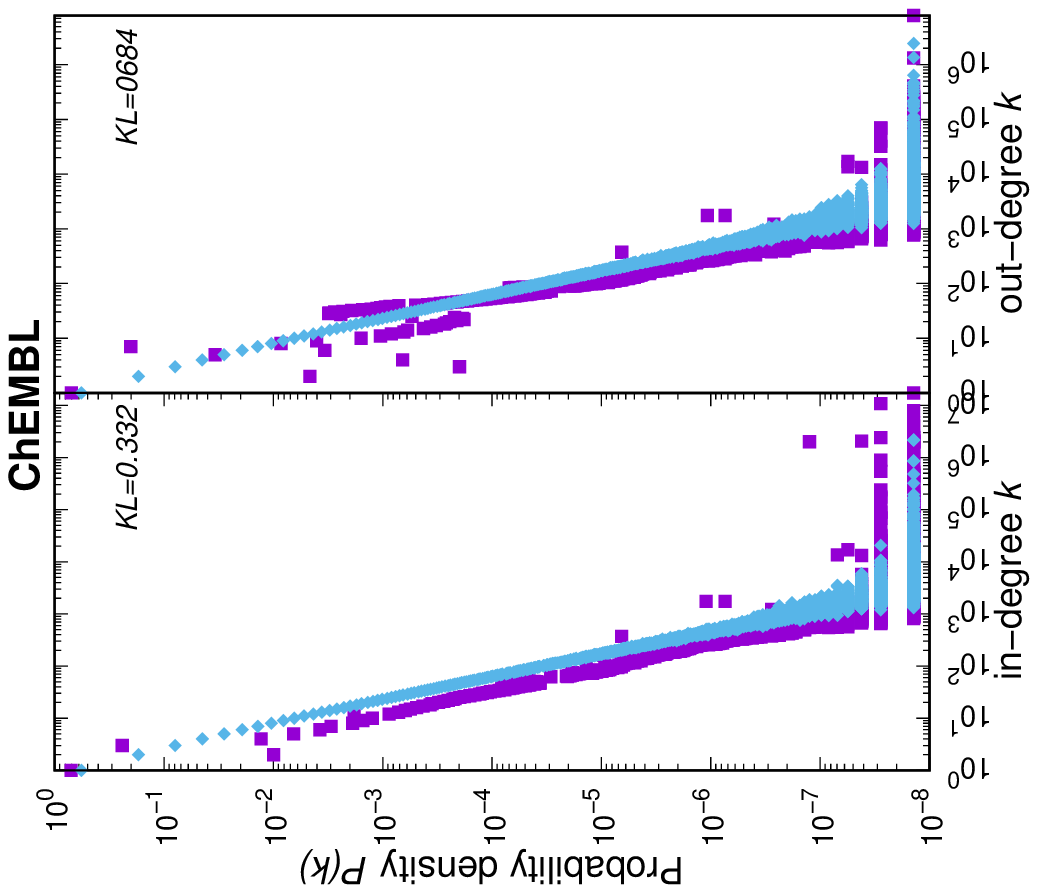}
\includegraphics[angle=-90, scale=0.33, keepaspectratio]{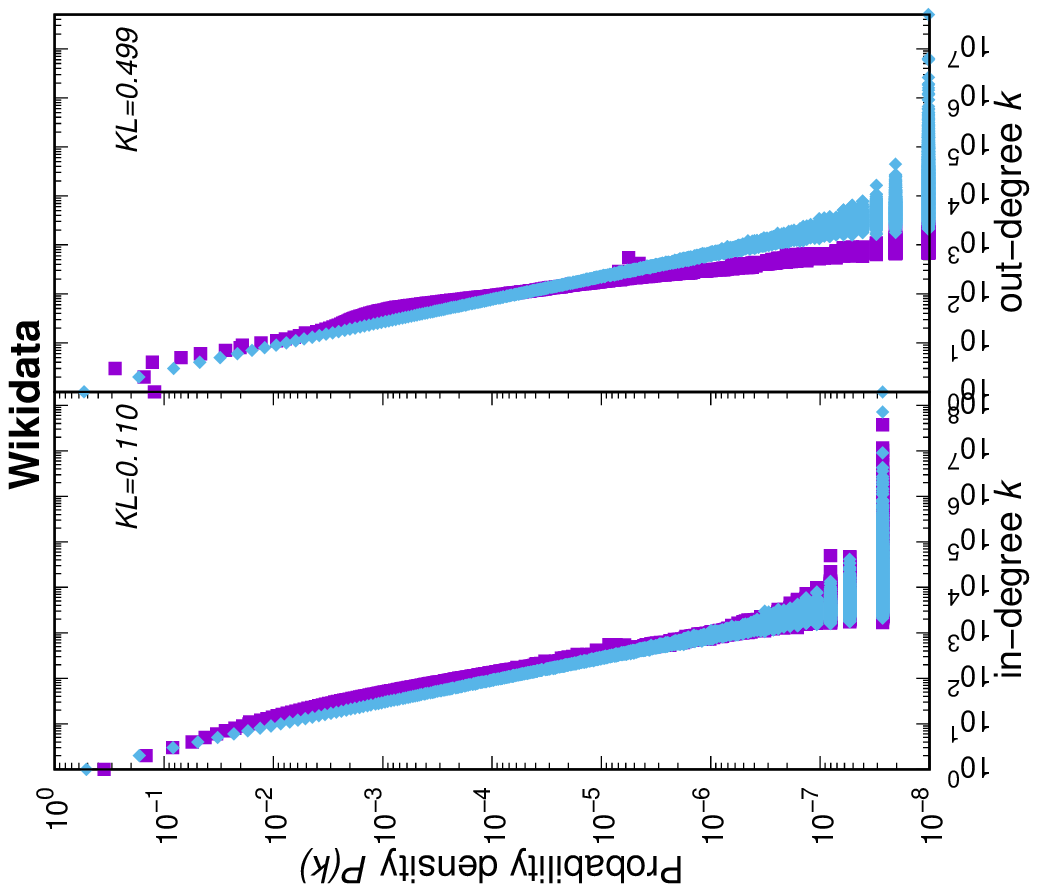}

\bigskip
(b) Bollob\'as model

\includegraphics[angle=-90, scale=0.33, keepaspectratio]{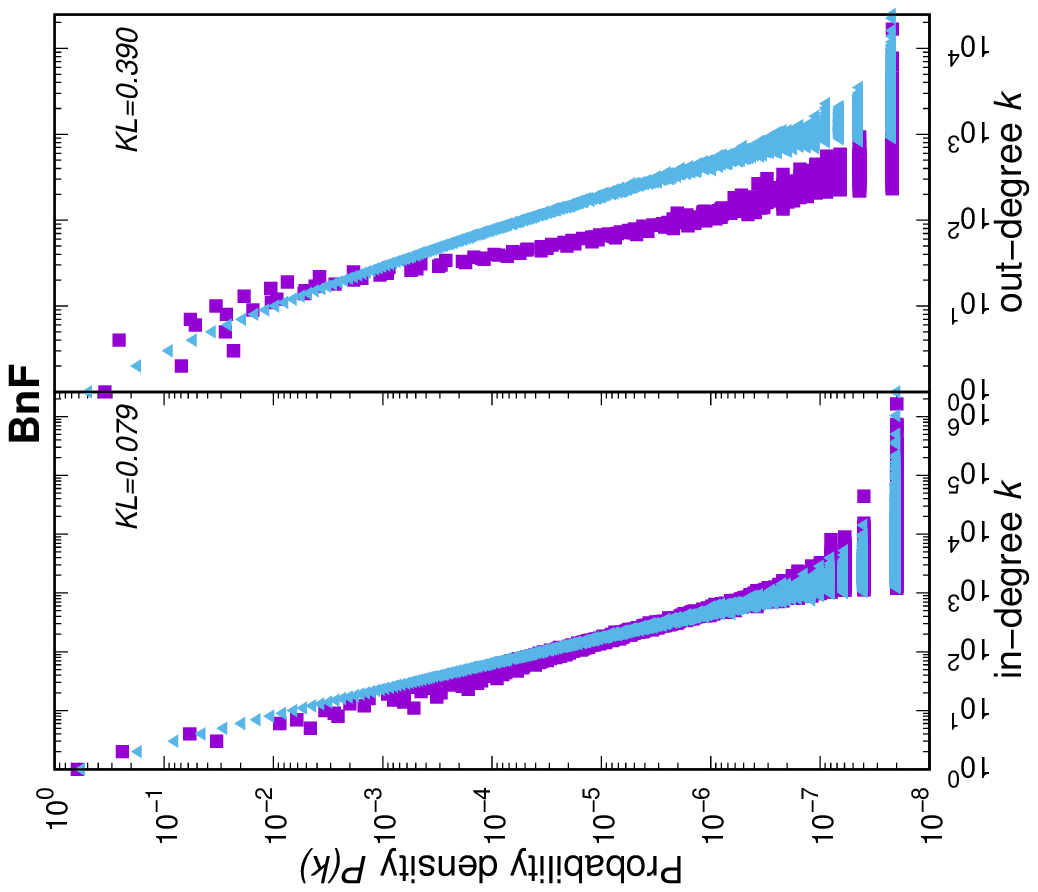}
\includegraphics[angle=-90, scale=0.33, keepaspectratio]{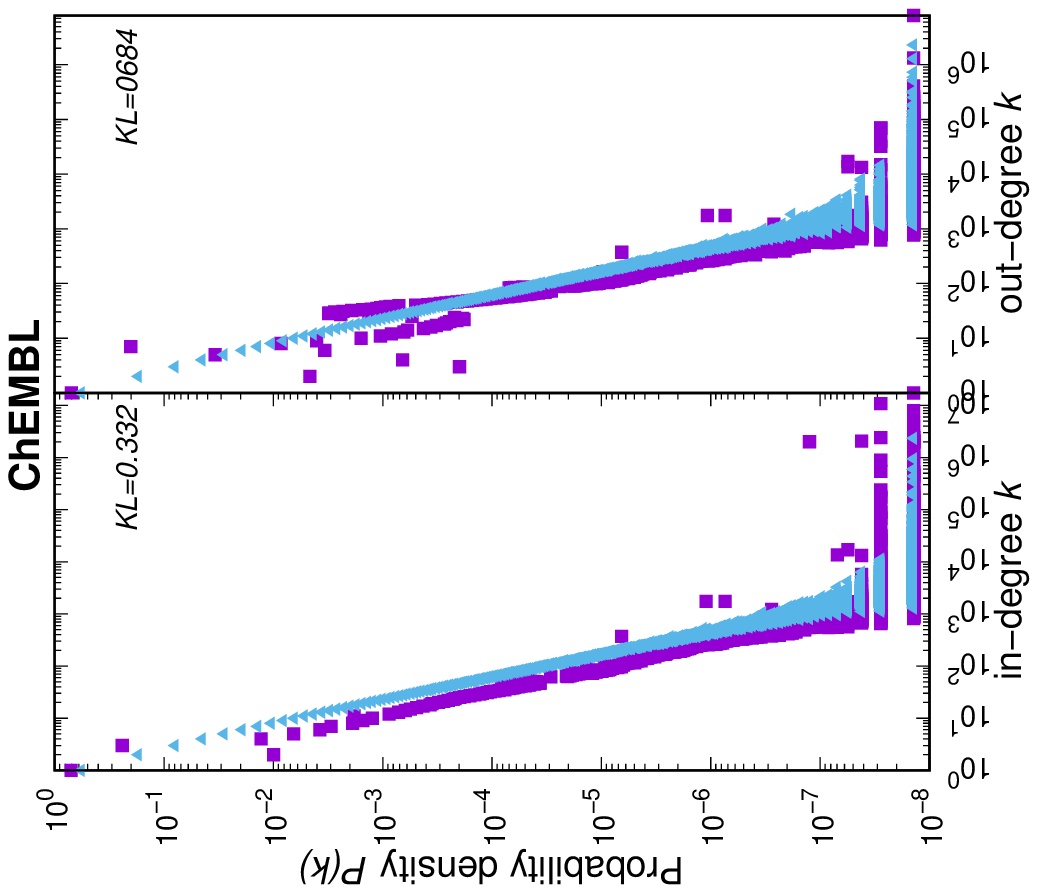}
\includegraphics[angle=-90, scale=0.33, keepaspectratio]{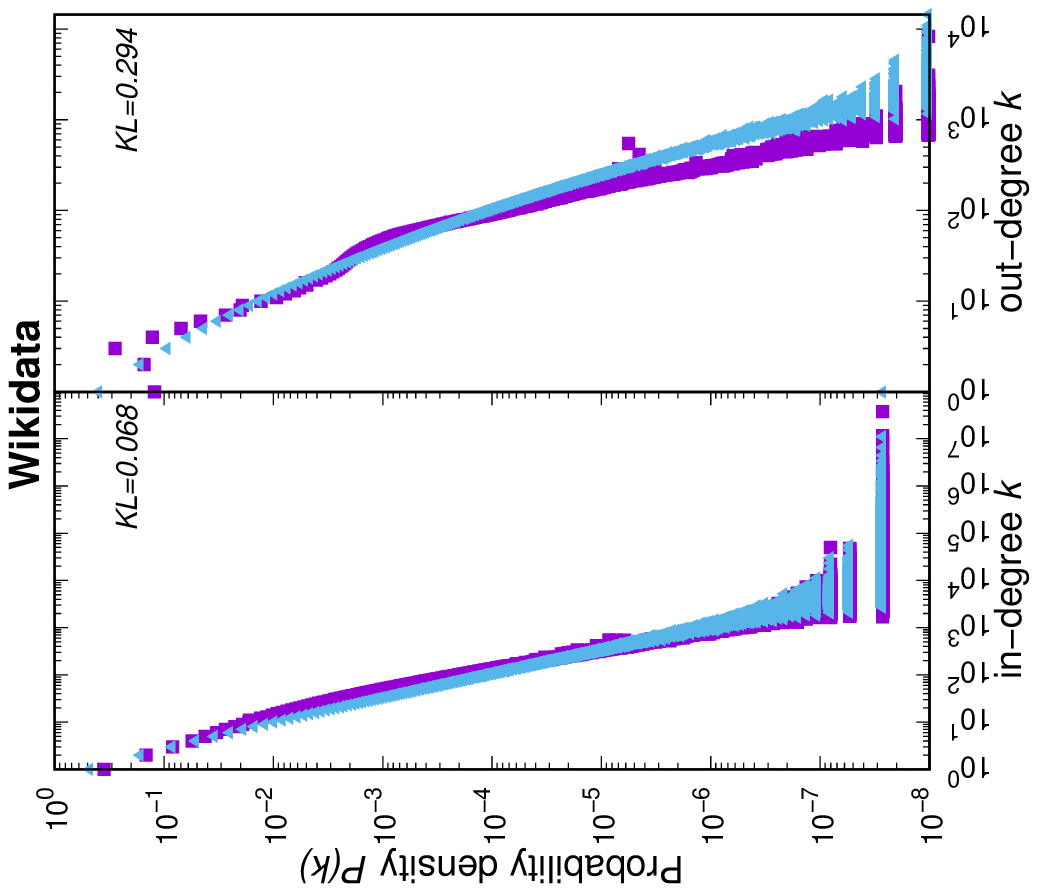}

}
}
\caption{\emph{Comparison of real-world data ({\tiny \textcolor{violet}{\ding{110}}}) and data generated by Barabási-Albert model ({\tiny \textcolor{cyan}{\ding{117}}}) and Bollob\'as model ({\tiny \textcolor{cyan}{\ding{115}}})  for BnF, ChEMBL and Wikidata.}
{ \small
Obviously, these two simplex models have a total connectivity $P(k)$ that always decreases with $k$, making it impossible to reproduce real data.
}
}
\label{fig:CompKGs}
\end{center}
\end{figure}


\clearpage


\section*{Supplementary materials}


\renewcommand{\theequation}{S\arabic{equation}}
\setcounter{equation}{0}


\subsection*{Materials and Methods}

\subsubsection*{Data sources and experimental setting}
\begin{sloppypar}
For the study of our generative model, we rely on three main knowledge graphs of the literature chosen for their large volume and for their diversity concerning their field (cultural heritage, chemistry, encyclopedia) and their curators (public service employees, academic researchers, general public).
\begin{itemize}
\item BnF assembles data from the various databases and catalogs of the French national library (Bibliothèque nationale de France, BnF) to provide federated access by authors, works, themes, places and dates. We used the dump (5.00 GB, gz compressed) downloaded on 05/29/2022 from the Web API of BnF \cite{bnfapi} in N-triple format that serializes Resource Description Framework (RDF) data.
\item ChEMBL is a manually curated chemical database of bioactive molecules with drug-like properties. Such data enables researchers to identify compounds for potential therapeutic targets, to investigate the available SAR or phenotypic data for a target and to identify the potential effects of specific chemotypes. 
%
We used the dump CHEMBL-RDF 28.0 \cite{chemblrdf} (2.06 GB, gz compressed) in Terse RDF Triple Language (Turtle).
\item
Wikidata is a free and collaborative knowledge graph that can be read and edited by both humans and machines 
(e.g., Wikidata acts as central storage for Wikipedia). We used the latest truthy dump (30.8GB bz2 compressed) downloaded on 04/29/2022 \cite{wikidatadump}  in NT format.
\end{itemize}
\end{sloppypar}

All experiments were performed on a 2.5 GHz Xeon processor with the Linux operating system and 32 GB of RAM.
The entire source code (the computation of the statistics and the generation of the knowledge graphs) is implemented in Java and it is available on the following git repository: \url{https://github.com/asoulet/superficiality}. 
Note that to facilitate scaling, we serialized the fact generation for each relationship in order to reduce memory consumption. For each relationship processed separately, this allowed us to use a binary search tree for sampling entities very efficiently despite a non-linear preferential attachment \cite{atwood2014efficient}. The in-depth description of these optimization techniques is out of  the scope of this article.

\begin{table}
\begin{center}
\begin{tabular}{lcrrrcr}
\hline
KG & Degree & \EntCard &	\KGCard &	$k_{\max}$ &	\newProb	& \RelCard \\
\hline
\hline
BnF	& IN & 50,213,129	& 211,066,099 &	1,639,696	& 0.610 &	647 \\
\cline{2-7}
 & OUT & 45,760,173 & 211,066,099 & 16,771 & 0.296 & 647 \\
\hline
ChEMBL & IN & 71,795,571 &	293,659,143 &	17,276,356 &	0.997 &	50 \\
\cline{2-7}
 & OUT & 71,697,849 &	293,659,143 &	7,927,537 &	0.378 &	50 \\
\hline
Wikidata & IN & 37,256,044 &	675,226,687 &	37,656,116 &	0.797 &	1,499 \\
\cline{2-7}
 & OUT & 98,005,966 & 675,226,687 & 8,322 & 0.284 & 1,499 \\
\hline
\end{tabular}
\caption{Main characteristics of preprocessed knowledge graphs}
\label{tab:Characteristics}
\end{center}
\end{table}

\begin{table}
\begin{center}
\begin{tabular}{crrrcr}
\hline
Year & \EntCard &	\KGCard &	$k_{\max}$ &	\newProb	& \RelCard \\
\hline
\hline
\multicolumn{6}{c}{IN} \\
\hline
2016 & 3,452,703 & 49,384,937& 2,776,706 & 0.716 & 717 \\
2017 & 5,577,565 & 81,321,093& 3,434,706 & 0.714 & 874 \\
2018 & 11,277,191 & 170,745,741& 14,340,630 & 0.785 & 1,010 \\
2019 & 21,267,757 & 352,323,629& 20,685,029 & 0.832 & 1,169 \\
2020 & 27,113,924 & 508,787,276& 36,624,592 & 0.782 & 1,365 \\
2021 & 28,514,756 & 525,822,216& 37,130,480 & 0.781 & 1,407 \\
2022 & 37,256,044 & 675,226,687& 37,656,116 & 0.797 & 1,499 \\
\hline
\hline
\multicolumn{6}{c}{OUT} \\
\hline
2016 & 15,448,893 & 49,384,937& 877 & 0.348 & 717 \\
2017 & 23,583,398 & 81,321,093& 1,526 & 0.340 & 874 \\
2018 & 43,847,957 & 170,745,741& 1,568 & 0.357 & 1,010 \\
2019 & 54,070,461 & 352,323,629& 8,321 & 0.325 & 1,169 \\
2020 & 90,393,612 & 508,787,276& 8,320 & 0.303 & 1,365 \\
2021 & 92,759,377 & 525,822,216& 8,320 & 0.300 & 1,407 \\
2022 & 98,005,966 & 675,226,687& 8,322 & 0.284 & 1,499 \\
\hline
\end{tabular}
\caption{Longitudinal study of Wikidata}
\label{tab:Longitudinal}
\end{center}
\end{table}

\subsubsection*{Preprocessing of entity knowledge graphs}

We filtered each dump to remove literals and external entities because our model aims at understanding the internal topology of the entity belonging to a given knowledge graph. In the same way that the study of the topology of the Web considers only the pages and their links (i.e., the content of the pages like text and images is ignored).
Of course, only the nodes of the knowledge graph corresponding to entities have been kept. Literal values such as dates, strings or images have therefore been removed.
Besides, for the BnF and Wikidata knowledge graphs, many relationships link subjects to entities belonging to other external knowledge graphs. For focusing on one graph at a time, we only consider the entities whose Uniform Resource Identifier (URI) is prefixed by \url{http://data.bnf.fr} or \url{http://www.wikidata.org/} for the BnF or Wikidata knowledge graph respectively.

All the results presented in this article concern entity knowledge graphs that are simply referred to as knowledge graphs. Table~\ref{tab:Characteristics} provides the main statistics of these knowledge graphs namely the number of entities \EntCard, the number of facts \KGCard, the maximum degree $k_{\max}$, the superficiality \newProb and the number of relationships \RelCard. For each knowledge graph, the information on the incoming degrees (corresponding to the objects) are distinguished from those on the outgoing degrees (corresponding to the subjects). The numbers of relationships and facts are common. For instance, in the BnF knowledge graph, there are  211M facts with 45.8M distinct entities as subject linked to 50.2M distinct entities as object.
To analyze the evolution of superficiality, Table~\ref{tab:Longitudinal} shows the Wikidata characteristics for each year since 2016 using the same approach as the preprocessing described for 2022. The dumps were downloaded from the Internet Archive \cite{internetarchive}.

\subsubsection*{Computation of statistics and parametrization of the generative model}

All statistics were computed by making ten passes on the dumps (five passes for in-degrees and five passes for out-degrees) in order not to overload the memory.
The first pass consists in computing the in/out-degree of each entity of the graph to build the ground truth distribution. We also get the number of entities \EntCard and the number of facts \KGCard reported in Table~\ref{tab:Characteristics}.
The other four passes are used to compute the in/out-degree ground truth distributions per relationship.
More precisely, the relationships are then divided into four groups $(R_i)_{i \in \{1,\dots,4\}}$. For each group $R_i$, we repeated a pass on the data that computes the number of entities $\EntCard_{\rel}$ and the number of facts $\KGCard_{\rel}$ for each relationship $\rel \in R_i$. This information allows us to estimate the relationship probability $\propProb = \KGCard_{\rel} / \KGCard$ and the attachment probability $\attachProb = 1 - \EntCard_{\rel} / \KGCard_{\rel}$.
As explained before, the superficiality parameter \newProb is then calculated directly:
$$\newProb =  \frac{\EntCard}{\sum_{\rel \in \Rel} \propProb \times (1 - \attachProb) \times \KGCard}$$

All the previous parameters are naturally derived from knowledge graphs. It remains however to parameterize the exponents \attachExp for each relationship \rel. For this purpose, we benefit from the rate at which an entity acquires facts given below:

\begin{equation}
\label{equ:Rate}
\frac{\partial \degree}{\partial t} = \attachProb \times \frac{\degree^{\attachExp}}{Z(t)}
\end{equation}

We can decompose $Z(t)$ by distinguishing the $z$ entities depicted by the relationship \rel with their number of facts $(Z_i)_{i \in \{1,\dots,z\}}$: $Z(t) = Z_1^{\attachExp} + \dots + Z_z^{\attachExp}$ where $\mean{z} = (1 - \attachProb) \times t$. $Z(t)$ is thus maximized if the $t$ facts are equally distributed among the $z$ entities. Consequently, we get a tight upper bound: $Z(t) \le \mean{z} \left( {t}/{\mean{z}} \right)^{\attachExp}$, which gives the following simplification: 
\begin{equation}
\label{equ:Constant}
Z(t) \le \frac{t}{(1 - \attachProb)^{\attachExp - 1}}
\end{equation}
Injecting this equation into Equation~\ref{equ:Rate}, we obtain the following inequality for $\attachExp < 1$ by rewriting the differential equation:
$$\int_1^{\degree_e} \frac{1}{\degree^{\attachExp}} \partial \degree \ge \attachProb \times (1 - \attachProb)^{\attachExp - 1} \times \int_{t_e}^{t} \frac{1}{t} \partial t$$
For an entity \ent added at time $t_\ent$ in the knowledge graph, the above equation gives its degree $\degree_\ent$ at time $t \ge t_\ent$. Integrating both sides and rearranging the terms, we get:
$$\degree_e \ge \left(\attachProb (1-\attachProb)^{\attachExp - 1} (1-\attachExp) \ln \frac{t}{t_e}+ 1\right)^{1/(1-\attachExp)}$$

The average maximum degree \maxDegree for a relationship can then be approximated by taking $t_e = 1$:

\begin{equation}
\label{equ:MaxDegree}
\mean{\maxDegree} \sim \left(\attachProb (1-\attachProb)^{\attachExp - 1} (1-\attachExp) \ln t + 1\right)^{1/(1-\attachExp)} 
\end{equation}

In practice, we determine the parameter \attachExp by searching for the value of this expression that is closest to the maximum degree measured in the data using a dichotomic search.
To evaluate the quality of our parameterization method, we generated synthetic relationships as ground truth by varying the parameters \attachExp and \attachProb between 0 and 1. We then applied our parameterization method for setting \attachExp and we compared the distribution from the found parameter with that of the ground truth. The Kullback-Leibler divergence is on average 0.004 (with a standard deviation of 0.003) and the maximum value is 0.032 (for  \attachExp = 0.05 and \attachProb = 0.35). These very low values for Kullback-Leibler divergence indicate that the parameterization method is satisfactory.



\subsubsection*{Ablation study}

Our method is based on a multiplex generative model (denoted by ``Multi.'') where each layer has a distinct preferential attachment mechanism parameterized between 0 and 1 (denoted by ``Param.''). Table~\ref{tab:Results} reports an ablation study using only one layer (denoted by ``Simp.'') or forcing a linear preferential attachment (denoted by ``Linear''). For simplex graphs, all links are merged into a single relationship. It means that the two variants are comparable to the generative models of Barab\'asi-Albert for ``Simp.+Linear'' \cite{barabasi1999emergence} and Bollob\'as for  ``Simp.+Param'' \cite{bollobas2003directed}. First, we observe that the parameterization of the preferential attachment is always beneficial (see the improvement of ``Multi.+Param.'' over ``Multi.+Linear''). Even if the gain is sometimes non-existent or weak for some configurations like BnF, it is essential for others like Wikidata. Second, the use of a multiplex model is relevant except for the case of ChEMBL out-degree connectivity. This issue stems from the poor modeling of $P(k_e(t)=1)$ that has a predominant weight with the Kullback-Leibler divergence. Of course, the simplex approaches do not take into account the oscillations contrary to our generative model. Finally, on these three knowledge graphs, it is clear that our model is the most satisfactory.

\begin{table}
\begin{center}
{\small
\begin{tabular}{lccccc}
\hline
KG & Type & Multi.+Param. & Multi.+Linear & Simp.+Param. & Simp.+Linear \\
\hline
\hline
BnF & IN &	\textbf{0.053} & \textbf{0.053} & 0.079 & 0.078 \\
\cline{2-6}
& OUT &	\textbf{0.273} & 0.278 & 0.390 & 0.485 \\
\hline
ChEMBL & IN &	\textbf{0.294} & 0.399 & 0.332 & 0.332 \\
\cline{2-6}
& OUT &	1.526 & 1.568 & \textbf{0.684} & \textbf{0.684} \\

\hline
Wikidata & IN &	\textbf{0.041} & 0.152 & 0.068 & 0.110 \\
\cline{2-6}
& OUT &	\textbf{0.057} & 0.117 & 0.294 & 0.499 \\
\hline
\end{tabular}
}
\caption{Generative model comparison of Kullback-Leibler divergences}
\label{tab:Results}
\end{center}
\end{table}


\subsection*{Supplementary Text}






\subsubsection*{Main results}
The number of distinct relationships that are attached to some entity is a main parameter to measure the knowledge relative to this entity.
Formula \ref{equ:Probability} in the main text precisely describes the proportion of entities that are attached to a fixed number of relationships. This section is devoted to the rigorous proof of this main result.

Let $\graph(t)$ be the knowledge graph obtained after $t$ steps
of the process. At each step, a relationship $\rel$ is chosen with probability $\propProb$ and is attached to an entity according to the following strategy: (a) by using the preferential attachment with probability $\attachProb$, (b) by creating a new entity with probability $(1-\attachProb)\newProb$ and (c) by  uniformly choosing a new existing entity with probability $(1-\attachProb)(1-\newProb)$. 

If we assume that in the initial graph $\graph(0)$, all relationships are attached to at least one entity, then cases (a) and (b) are always possible. However, case (c) may not occur when a relationship is already attached to all entities. We will show later that this event has a low probability and we will call generic (resp. exceptional), a step where the three cases are possible (resp. not possible).

Some basic results can be quickly established on the graph $\graph(t)$. First, a new fact is added at each step and the number of facts in $G(t)$ is exactly $G(0)+t$. Let $\degree_\rel(t)$ denote the number of facts that involve relationship $\rel$. The probability to add a new fact with relationship $\rel$ is $\propProb$, independently from the past, so that  $\degree_\rel(t)-\degree_\rel(0)$ follows a binomial law with parameters $(t,\propProb)$. In particular, the mean and variance are linear in $t$ and
$$
\mean{\degree_\rel(t)}=\degree_\rel(0)+\propProb t,\qquad
\variance{\degree_\rel(t)}=\propProb(1-\propProb) t.
$$
The parameter $\degree_\rel(t)$ counts the total number of entities attached to $\rel$, counted with their multiplicities (an entity may be attached several times to a same relationship). Consider now  $\EntCard_\rel(t)$,  the number of distinct entities attached to some relationship $\rel$ at time $t$. Since the probability to attach a new entity to $\rel$ at each step is $\cc_\rel=\propProb(1-\attachProb)$, independently from the past, the quantity $\EntCard_\rel(t)-\EntCard_\rel(0)$ also follows a binomial law with parameters $(t,\cc_\rel)$. The mean and the variance of $\EntCard_\rel$ then satisfy
$$
\mean{\EntCard_\rel(t)}=\EntCard_\rel(0)+\cc_\rel t,\qquad
\variance{\EntCard_\rel(t)}= \cc_\rel(1-\cc_\rel) t
$$
and with high probability, $\EntCard_\rel(t)$ is asymptotically close to its mean. Precisely, Chernoff bounds give the following large deviation property for all fixed $\epsilon>0$,
$$
\Prob{\left| \EntCard_\rel(t)-\cc_\rel t\right|>\sqrt{\epsilon t\ln t}}=\mathcal{O}
\left(t^{-2\epsilon}\right),
$$
with a constant in $\mathcal{O}$ independent of $\epsilon$.
Intuitively when the process mainly contains generic steps, the probability to add a new entity is  $\aa=\sum_{\rel=1}^\RelCard \propProb(1-\attachProb)\newProb$ (at each step) and the expected number of entities, denoted by $\EntCard(t)$, should be asymptotically close to $\aa t$. 
The following result gives more details.

\noindent {\bf Result 1. } \emph{Suppose $\aa>\max_{\rel\in\Rel}\cc_\rel$. The total number of entities, denoted by $\EntCard(t)$, has linear asymptotic mean and variance (with respect to $t$) and satisfies a central limit theorem, 
    $$
    \mean{\EntCard(t)}\mathop{\sim}_{t\to\infty} \aa t,\quad \variance{\EntCard(t)} \mathop{\sim}_{t\to\infty} \aa(1-\aa)t\quad\mbox{and}\quad
    \frac{\EntCard(t)-\aa t}{\sqrt{\aa(1-\aa)t}}\mathop{\to}_{t\to\infty} \mathcal{N}(0,1).
    $$
There also exists some constant $\bb>0$ such that for all fixed $\epsilon>0$,
$$
\Prob{\left| \EntCard(t)-\aa t\right|>\sqrt{\epsilon t\ln t}}=\mathcal{O}
\left(t^{-\bb\epsilon}\right),
$$
with a constant in $\mathcal{O}$ independent of $\epsilon$.
    }

A step is generic as soon as all the relationships are not attached to all the entities, or equivalently $\EntCard(t)>\EntCard_\rel(t)$ for all relationships $\rel$.  Asymptotically, this condition is equivalent to the hypothesis $\aa>\max_{\rel\in\Rel}\cc_\rel$ in Result 1. The proof of Result 1 is not immediate, but it can be obtained using generating functions and the Analytic Combinatorics approach \cite{FS09}.  

For $i\in \mathbb{N}^\star$, let $\EntCardi_i(t)$ be the total number of entities being attached to exactly $i$ relationships. The study of the parameters $\EntCardi_i(t)$ can be done in the general model where all the $\propProb$ and $\attachProb$ are distinct but for the sake of simplicity, we suppose that they are constant and equal to $\propProbC$ and  $\attachProbC$. We now state our second main result.

\noindent {\bf Result 2. } \emph{
Fix $i\in\mathbb{N}^\star$. The mean number of entities being attached to exactly $i$ distinct relationships is asymptotically linear in $t$ and satisfies
    $$
    \mean{\EntCardi_i(t)} \mathop{\sim}_{t\to\infty} \frac{K_1K_2\dots K_{i-1}}{(1+K_1)\dots (1+K_i)}\newProb(1-\attachProbC)t\quad\mbox{with}\quad K_i = \frac{1 - \newProb}{\RelCard \newProb - {1}}\left(\RelCard-{i}\right).
    $$
In addition, the mean value of the proportion ${\EntCardi_i(t)}/{\EntCard(t)}$ of entities that are described by few relationships (misdescribed entities) is
$$
\mean{\frac{\EntCardi_i(t)}{\EntCard(t)}} \mathop{\sim}_{t\to\infty} \frac{K_1K_2\dots K_{i-1}}{(1+K_1)\dots (1+K_i)}.
$$
}
Note that $K_n=0$. The second asymptotic is a direct consequence of the first one and the fact that $\EntCard(t)$ is concentrated around its mean and is lower bounded by $m(0)$ (that we can suppose to be strictly positive for convenience). Remark that  $\mean{{\EntCardi_i(t)}/{\EntCard(t)}}$
is the mathematical version of the proportion $P(\rel_{\ent}(t) \le i)$ used in the main text.
When $n$ and $t$ tends to infinity, $\mean{\EntCardi_i(t)/\EntCard(t)}$ tends to a geometric law, precisely
$$
\lim_{n\to\infty} \lim_{t\to\infty} \mean{\frac{\EntCardi_i(t)}{\EntCard(t)}}=\newProb(1-\newProb)^{i-1}\quad\mbox{or}\quad
\lim_{n\to\infty}\lim_{t\to\infty} \mean{\frac{{\EntCardi_1(t)+\dots+\EntCardi_i(t)}}{{\EntCard(t)}}}=1- (1-\newProb)^{i}
.
$$

\subsubsection*{Result 1: The proof of the average number of entities}
The Analytic Combinatorics method aims at describing the generic behavior of algorithms or data structures when their size becomes large. Developed since the 1980s around Philippe Flajolet and Robert Sedgewick, this method relies on the formal manipulation of generating functions and their analytical properties to elucidate the asymptotic behavior of parameters of interest. A detailed presentation of this approach is given in the book~\cite{FS09}. In this section, this method is used to perform a precise analysis of the parameter $\EntCard(t)$.

For $t,k\geq 0$, let $p_{t,k}=\Prob{\EntCard(t)=k}$ be the probability distribution of $\EntCard(t)$ and consider $q_{t,\rel,k}=\Prob{m(t)=m_\rel(t)=k}$ the probability to be in an exceptional case due to relation $\rel$.
Basic probabilistic computations give the recurrence formula
\begin{eqnarray}
p_{t+1,k}&=&(1-a)p_{t,k}+ap_{t,k-1}+\sum_{\rel=1}^n \rho_\rel (1-\beta_\rel)(1-\sigma) \left(q_{t,\rel,k-1}-q_{t,\rel,k}\right)\label{eq:1}\\
p_{0,k}&=&\mathbf{1}[k=m(0)].\nonumber
\end{eqnarray}
Intuitively, $ap_{t,k-1}$ is the \emph{generic} probability to generate a new entity when $k-1$ entities exist whereas $(1-a)p_{t,k}$ is the \emph{generic} probability not to generate a new entity when $k$ entities exist. The last sum corresponds to the exceptional cases.
The bivariate generating functions $P(z,u)$ and $Q_r(z,u)$ relative to the the probabilities
$p_{t,k}$ and $q_{t,\rel,k}$ are defined by
$$
P(z,u)=\sum_{t\geq 0}\sum_{k\geq m(0)}p_{t,k}z^tu^k\qquad\mbox{and}\qquad
Q_r(z,u)=\sum_{t\geq 0}\sum_{k\geq m(0)}q_{r,t,k}z^tu^k.
$$ 
Based on the previous recurrence relations, alternative expressions of the generating functions are derived so that
$$
P(z,u)=\frac{u^{m(0)}+(1-u)\displaystyle\sum_{\rel=1}^n \rho_\rel (1-\beta_\rel)(1-\sigma) Q_\rel(z,u) }{1-z(1-a+au)}.
$$
It is well known in complex analysis that the asymptotic of the coefficients of  $P(z,u)$ is closely related to its singularities. 

Using the recurrence relation \eqref{eq:1}, the inequality $
q_{\rel,t,k}\leq \min\left(p_{t,k},\Prob{m_r(t)=k}\right)$, Chernoff bounds with the binomial law of $m_\rel(t)-m_\rel(0)$ and the fact that $a>\max_{\rel\in\Rel} c_\rel$, we prove that there exists some $\eta_\rel>0$  such that 
$$
q_{\rel,t,k}=O\left(e^{-\eta_\rel t}\right)\quad \mbox{uniformly for $k=m_\rel(0)..m_\rel(0)+t$}
$$
the other coefficients being nul. In particular, the bivariate generating function  $Q_\rel(z,u)$
is analytic in $(z,u)\in B(0,e^{\eta_r/3})^2$.
Taking $\eta= \min_\rel \eta_\rel/3$, the numerator is analytic for  $(z,u)\in B(0,e^{\eta})^2$. Then there exists a complex neighborhood  $\mathcal{V}_1$ of $u=1$ such that  $P(z,u)$ is meromorphic on $B(0,e^{\eta})\times\mathcal{V}_1$ with a unique simple pole in $z=1/(1-a+au)$.
This is the meromorphic scheme of Theorem IX.8 in\cite{FS09} which gives the asymptotics of the expectation, variance and the Gaussian  limit law of Result 1. Theorem IX.15 also applies and entails the large deviation inequality.

\subsubsection*{Result 2: The proof of the proportion of misdescribed entities}
In this section, we suppose that the parameters $\attachProb$ and $\propProb$ are constant and equal to $\attachProbC$ and $\propProbC$ respectively. In this context, the constants $c_\rel$ does not depend on $\rel$ and we write $c$ for $c=c_\rel=(1-\attachProbC)/\RelCard$.
With high  probability,  the number of entities $\EntCard(t)$ is close to $at$ and the number of distinct entities attached to the relationships $\rel$ is close to $c t$. Let $\mathcal{G}(t)$ be the \emph{generic event} 
$$
\mathcal{G}(t)=\left(\bigcap_{\rel\in\Rel}\left[\left|\EntCard_\rel(t)-c t\right|<\sqrt{\epsilon t\ln t}\right] \right)\cap\left[\left|\EntCard(t)-at\right|<\sqrt{\epsilon t\ln t}\right].
$$
Result 1 and the Chernoff bounds applied to the binomial laws of the $\EntCard_\rel$ entail that the probability of $\mathcal{G}(t)$ is close to $1$. Precisely one has with a sufficiently large $\epsilon$, 
$$
\Prob{\mathcal{G}(t)} = 1+{\cal O}(t^{-3}).
$$
Asymptotically with high probability, the process is then always in the generic case.
The parameter $\EntCardi_{i+1}$ increases by $1$ at time $t+1$ if there exists an entity $e$,  attached to exactly $i$ distinct relationships at time $t$, that is attached to a new relationship at time $t+1$. For such entity  $e$, this event occurs with probability equivalent to $\frac{(n-i)}{n}(1-\attachProb)(1-\newProb)\frac{1}{t(a-c)}=\frac{K_i}{t}$ as $t$ tends to $\infty$. Indeed, $\frac{(n-i)}{n}$ is the probability to choose a new relationship for $e$ and then $(1-\attachProbC)(1-\newProb)\frac{1}{t(a-c)}$ is the asymptotic probability to perform an attachment to $e$ (case (c)) in the generic case. 
Note that by symmetry at time $t+1$, $\EntCardi_{i}$ decreases by $1$ as soon as $\EntCardi_{i+1}$ increases by $1$ (with probability equivalent to $K_i/t$).
Since there exist $\EntCardi_i(t)$ entities at time $t$ attached to exactly $i$ relationships and taking into account the non-generic case, we obtain the recurrence formula
\begin{eqnarray}
\mean{\EntCardi_i(t+1)}&=&\mean{\EntCardi_i(t)}+\frac{K_{i-1}}{t}\mean{\EntCardi_{i-1}(t)}- \frac{K_{i}}{t} \mean{\EntCardi_{i}(t)}+O\left(t^{-1/2}\sqrt{\ln t}\right)\nonumber\\
&=&\left(1-\frac{K_{i}}{t}\right)\mean{\EntCardi_{i}(t)}+\frac{K_{i-1}}{t}\mean{\EntCardi_{i-1}(t)}+O\left(t^{-1/2}\sqrt{\ln t}\right).\label{eq:x1}    
\end{eqnarray}
For $i=1$, the probability that  $\EntCardi_1(t)$ increases is the probability to create a new entity in the generic case which is $(1-\attachProbC)\newProb$ so that
\begin{eqnarray}
\mean{\EntCardi_1(t+1)}&=&\mean{\EntCardi_1(t)}+(1-\attachProbC)\newProb- \frac{K_{1}}{t} \mean{\EntCardi_{1}(t)}+O\left(t^{-1/2}\sqrt{\ln t}\right)\nonumber\\
&=&\left(1-\frac{K_{1}}{t}\right)\mean{\EntCardi_{1}(t)}+(1-\attachProbC)\newProb+O\left(t^{-1/2}\sqrt{\ln t}\right)\label{eq:xi}.
\end{eqnarray}
To conclude, we will apply the following lemma for each $\EntCardi_i$.

\medskip

\noindent {\bf Lemma 1.} \emph{Fix $a$ and $b$ two strictly positive constants. Consider the real sequence $(x_t)_{t\geq 1}$ satisfying the following recurrence
$$
x_{t+1}=\left(1-\frac{a}{t}\right)x_t+b+\epsilon_t\quad\mbox{with}\quad \epsilon_t=\mathcal{O}\left(t^{-1/2}\sqrt{\ln t}\right).
$$
Then \qquad 
$
\displaystyle x_t = \frac{b}{1+a}t + \mathcal{O}\left(t^{-1/2}\sqrt{\ln t}\right).
$
}

\medskip

\noindent 
The previous lemma applied with $a=K_1$ and $b=(1-\attachProbC)\newProb$ gives the announced result for $\EntCardi_1(t)$,
$$
\EntCardi_1(t)=\frac{1}{1+K_1}(1-\attachProbC)\newProb t + \mathcal{O}\left(t^{-1/2}\sqrt{\ln t}\right).
$$
It is now possible to insert the asymptotic of $\EntCardi_1(t)$ in Equation \eqref{eq:xi} for $\EntCardi_2$ so that
$$
\mean{\EntCardi_2(t+1)}=\left(1-\frac{K_{2}}{t}\right)\mean{\EntCardi_{2}(t)}+\frac{K_{1}}{1+K_1}(1-\attachProbC)\newProb     +O\left(t^{-1/2}\sqrt{\ln t}\right).    
$$
Once more, Lemma 1 applies and the asymptotic for $\EntCardi_2$ can be computed. The same process gives the result for $\EntCardi_3$, then $\EntCardi_4$ and so on.



\end{document}